\documentclass[conference]{IEEEtran}
\IEEEoverridecommandlockouts
\usepackage{amsmath,amssymb,amsfonts}
\usepackage{algorithmic}
\usepackage{graphicx}
\usepackage{textcomp}
\usepackage{multirow}
\usepackage{amsmath}
\usepackage{subfigure}
\usepackage{xcolor}
\def\BibTeX{{\rm B\kern-.05em{\sc i\kern-.025em b}\kern-.08em
    T\kern-.1667em\lower.7ex\hbox{E}\kern-.125emX}}
\usepackage[hidelinks]{hyperref}
\usepackage[sorting=none]{biblatex}
\hypersetup{
  colorlinks   = true, 
  urlcolor     = blue, 
  linkcolor    = blue, 
  citecolor   = blue 
}

\AtBeginBibliography{\small}

\begin{document}

\title{Optimizing ROI Benefits Vehicle ReID in ITS

\thanks{This work was supported by the Joint Transportation Research Program (JTRP), administered by the Indiana Department of Transportation and Purdue University, Grant SPR-4436.}
}

\author{%
  \IEEEauthorblockN{%
  Mei Qiu, Lauren Ann Christopher$^*$\thanks{$^*$Corresponding Author}, Lingxi Li, Stanley Chien, Yaobin Chen}%
{Purdue University Indianapolis, IN, USA}
}

\maketitle

\begin{abstract}
Vehicle re-identification (ReID) is a computer vision task that matches the same vehicle across different cameras or viewpoints in a surveillance system. This is crucial for Intelligent Transportation Systems (ITS), where the effectiveness is influenced by the regions from which vehicle images are cropped. 
This study explores whether optimal vehicle detection regions, guided by detection confidence scores, can enhance feature matching and ReID tasks. Using our framework with multiple Regions of Interest (ROIs) and lane-wise vehicle counts, we employed YOLOv8 for detection and DeepSORT for tracking across twelve Indiana Highway videos, including two pairs of videos from non-overlapping cameras. Tracked vehicle images were cropped from inside and outside the ROIs at five-frame intervals. Features were extracted using pre-trained models: ResNet50, ResNeXt50, Vision Transformer, and Swin-Transformer. Feature consistency was assessed through cosine similarity, information entropy, and clustering variance. Results showed that features from images cropped inside ROIs had higher mean cosine similarity values compared to those involving one image inside and one outside the ROIs. The most significant difference was observed during night conditions (0.7842 inside vs. 0.5 outside the ROI with Swin-Transformer) and in cross-camera scenarios (0.75 inside-inside vs. 0.52 inside-outside the ROI with Vision Transformer). Information entropy and clustering variance further supported that features in ROIs are more consistent. These findings suggest that strategically selected ROIs can enhance tracking performance and ReID accuracy in ITS.
\end{abstract}

\begin{IEEEkeywords}
Vehicle Detection, Vehicle Tracking, Region of Interests, Feature Matching, Feature Consistency, Vehicle Re-identification
\end{IEEEkeywords}

\section{Introduction}
Vehicle reidentification (ReID) in Intelligent Transportation Systems (ITS) remains a challenging task. Despite significant efforts to enhance performance on public benchmark datasets, there is a notable gap in methodologies for collecting custom vehicle ReID data under various ITS scenarios. Building ReID datasets by cropping from vehicle detection results has become a conventional practice in AI and smart city challenges \cite{tan2019multi, qian2020electricity, li2022multi}. Some research has set multiple Regions of Interest (ROIs) by positioning cameras for maximal clarity \cite{chang2019ai, ye2021robust} or by adaptively learning ROIs using detector confidence values, as objects detected in the ROI exhibit higher confidence values than those farther away or outside the ROI \cite{vasu2021vehicle, qiu2024intelligent}. Our previous work proposed automatic ROI learning on highway videos leveraging vehicle detection confidence scores \cite{qiu2024intelligent, lin2022identifying}, demonstrating that within these ROIs, vehicle tracking achieved more accurate and efficient results compared to regions outside these ROIs due to better feature representation.

In this work, we explore whether ROIs optimized based on vehicle detection confidence scores can improve vehicle re-identification in ITS, addressing questions that have not been previously investigated:
\begin{itemize}
    \item Can these ROIs, guided by vehicle detection results, enhance feature matching within a single camera view?
    \item Can these ROIs, guided by vehicle detection results, enhance feature matching across non-overlapping camera views?
\end{itemize}

\begin{figure*}
\centerline{\includegraphics[width=0.95\textwidth]{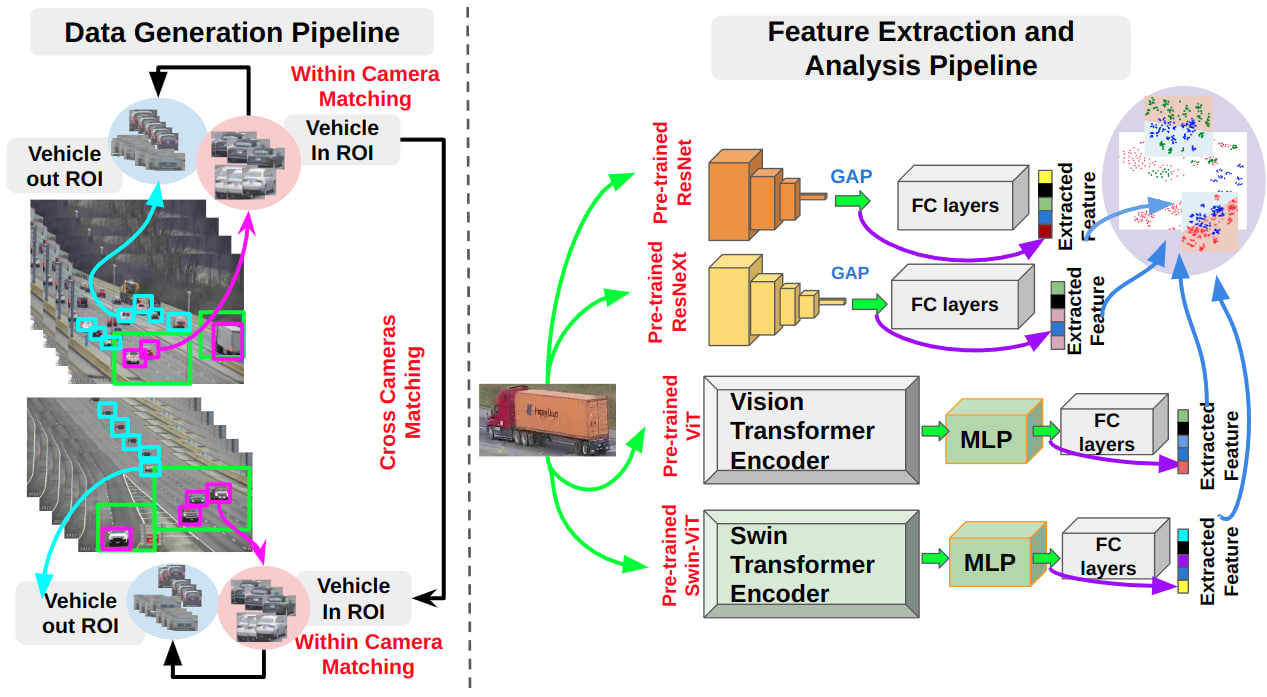}}
\vspace{-3mm}
\caption{\small \textit{\textbf{Data Generation Pipeline}:
   Within Camera Matching: Identifies and matches vehicles within the Region of Interest (ROI) and out of ROI in a single camera view.
   Cross Cameras Matching: Matches vehicles across different camera views, detecting them in both ROI and out of ROI, and matching vehicles in ROIs. \textbf{Feature Extraction and Analysis Pipeline}:
   Pre-trained ResNet and ResNeXt: Utilize Global Feature extraction after the layer of Average Pooling (GAP).
   Pre-trained Vision Transformer (ViT) and Swin Transformer: Utilize Global Feature extraction after the layer of Multi-Layer Perceptron (MLP). Cosine similarity, clustering, information entropy and T-SNE data analysis algorithms are used to analyze features' characteristics. \textbf{\color{red}{Best view in color}.}}}
\label{flowchart}
\vspace{-1mm}
\end{figure*}

\section{Problem Formulation}
To explore whether vehicle images cropped inside ROIs, based on vehicle detection confidence scores, enhance vehicle reidentification (ReID), we formulate this problem as a feature matching and consistency estimation task. ROIs on input frames are derived using vehicle detection confidence scores based on our previous work \cite{qiu2024intelligent}. Using a tracking method \cite{qiu2024real}, we crop images within these ROIs and split them into two classes: inside ROIs and outside ROIs. We hypothesize that images cropped inside ROIs will have stronger feature consistency, making them more suitable for the ReID task in ITS.

Formally, let $\mathcal{I}_\text{InROI}$ and $\mathcal{I}_\text{OutROI}$ denote the sets of images cropped inside and outside the ROIs, respectively. For a given feature extractor $\phi$, the feature sets are $\Phi_\text{InROI} = \{\phi(I) \mid I \in \mathcal{I}_\text{InROI}\}$ and $\Phi_\text{OutROI} = \{\phi(I) \mid I \in \mathcal{I}_\text{OutROI}\}$.

Feature consistency is measured using cosine similarity, $\text{Sim}(\cdot, \cdot)$. For each unique vehicle, let $n_i$ be the number of image pairs inside ROIs, and $\text{Sim}_{ii}(j)$ the cosine similarity of the $j$-th pair. The average $\text{Sim}_{ii}$ is computed over all $N$ unique vehicles. Similarly, let $n_o$ be the number of cross-ROI pairs, and $\text{Sim}_{io}(j)$ the cosine similarity of the $j$-th cross-ROI pair. The average intra-set similarity is given by:

\begin{equation}
\text{AvgSim}_\text{InROI} = \frac{1}{N} \sum_{i=1}^{N} \left( \frac{1}{n_i} \sum_{j=1}^{n_i} \text{Sim}_{ii}(j) \right)
\end{equation}

\begin{equation}
\text{AvgSim}_\text{OutROI} = \frac{1}{N} \sum_{i=1}^{N} \left( \frac{1}{n_o} \sum_{j=1}^{n_o} \text{Sim}_{io}(j) \right)
\end{equation}

We compare $\text{AvgSim}_\text{InROI}$ and $\text{AvgSim}_\text{OutROI}$ to assess feature consistency. If $\text{AvgSim}_\text{InROI} > \text{AvgSim}_\text{OutROI}$, we conclude that images inside ROIs have stronger feature consistency, making them more suitable for vehicle ReID in ITS.

\section{Methodology}
\smallskip
\noindent
\textbf{Data Collection.}
We start by analyzing videos to identify non-overlapping regions with optimal vehicle detection results, characterized by high confidence scores using our previous framework \cite{qiu2024intelligent}. Vehicle detection and tracking are performed across the entire frame \cite{qiu2024real}. Each tracked vehicle is labeled as "in ROI" if it enters the Region of Interest (ROI), or "out ROI" otherwise. For cross-camera scenarios, human labels of vehicles within ROI areas were created using a majority vote strategy. The data generation pipeline is shown in Fig. \ref{flowchart}.

\smallskip
\noindent
\textbf{Feature Extraction.}
We use four pre-trained models for feature extraction: ResNet50 \cite{he2016deep}, ResNeXt50 \cite{xie2017aggregated}, Vision Transformer \cite{dosovitskiy2020image}, and Swin-Transformer \cite{liu2021swin}. Features are extracted from these models before the classification layer. The feature extraction pipeline is illustrated in Fig. \ref{flowchart}.

\smallskip
\noindent
\textbf{Feature Matching and Consistency Analysis.}
To determine if vehicle images cropped inside ROIs improve ReID, we analyze feature consistency using cosine similarity. ROIs are derived from detection confidence scores \cite{qiu2024intelligent}, and images are cropped using a tracking method \cite{qiu2024real}. Cropped images are classified as inside or outside ROIs. We hypothesize that images inside ROIs have stronger feature consistency, making them more suitable for ReID.

Let $\mathcal{I}_\text{InROI}$ and $\mathcal{I}_\text{OutROI}$ denote sets of images cropped inside and outside ROIs, respectively. For a feature extractor $\phi$, we define $\Phi_\text{InROI} = \{\phi(I) \mid I \in \mathcal{I}_\text{InROI}\}$ and $\Phi_\text{OutROI} = \{\phi(I) \mid I \in \mathcal{I}_\text{OutROI}\}$.  We analyze cosine similarity for each vehicle, both inside and outside ROIs, using a t-test (\(p = 0.05\)). Additionally, t-SNE is used to visualize clustering differences. Define \( \text{sim}_{ii}(j) \) as the cosine similarity of the \( j \)-th vehicle pair inside the ROI, and \( \text{sim}_{io}(j) \) as the cosine similarity of the \( j \)-th pair with one vehicle inside and one outside the ROI.

The mean (\( \mu \)) and standard deviation (\( \sigma \)) of these similarities are computed as follows:

\textbf{For vehicles both from inside the ROI:}
\begin{equation}
\mu_{\text{inside}} = \frac{1}{N} \sum_{i=1}^{N} \left( \frac{1}{n_{\text{inside},i}} \sum_{j=1}^{n_{\text{inside},i}} \text{sim}_{ii}(j) \right),
\end{equation}
\begin{equation}
\sigma_{\text{inside}} = \sqrt{\frac{1}{N} \sum_{i=1}^{N} \left( \frac{1}{n_{\text{inside},i}} \sum_{j=1}^{n_{\text{inside},i}} \left( \text{sim}_{ii}(j) - \mu_{\text{inside}} \right)^2 \right)},
\end{equation}

\textbf{For pairs of vehicles, one from inside and one from outside the ROI:}
\begin{equation}
\mu_{\text{cross}} = \frac{1}{N} \sum_{i=1}^{N} \left( \frac{1}{n_{\text{cross},i}} \sum_{j=1}^{n_{\text{cross},i}} \text{sim}_{io}(j) \right),
\end{equation}
\begin{equation}
\sigma_{\text{cross}} = \sqrt{\frac{1}{N} \sum_{i=1}^{N} \left( \frac{1}{n_{\text{cross},i}} \sum_{j=1}^{n_{\text{cross},i}} \left( \text{sim}_{io}(j) - \mu_{\text{cross}} \right)^2 \right)},
\end{equation}

where \( n_{\text{inside},i} \) and \( n_{\text{cross},i} \) are the number of vehicle pairs for each unique vehicle in each setting, respectively.

\textbf{Hypothesis Testing:}
\begin{itemize}
    \item Null Hypothesis \( H_0 \): The mean cosine similarity for images cropped inside ROIs is equal to that for images involving one image inside and one outside the ROIs.
    \item Alternative Hypothesis \( H_1 \): The mean cosine similarity for images cropped within ROIs is higher than that for images involving one image inside and one outside the ROIs.
\end{itemize}
We use a t-test to compare the mean cosine similarities:
\begin{equation}
t = \frac{\mu_{\text{inside}} - \mu_{\text{cross}}}{\sqrt{\frac{\sigma_\text{inside}^2}{N} + \frac{\sigma_\text{cross}^2}{N}}}
\end{equation}
Assuming the sample sizes and standard deviations are sufficiently large to establish significance, we expect the t-test to reject the null hypothesis in favor of the alternative hypothesis, supporting our claim.

\smallskip
\noindent
\textbf{Feature Consistency Analysis with Information Entropy.}
Information entropy measures uncertainty or randomness in a dataset. To analyze feature consistency using entropy, we follow these steps:

\textbf{Discretize Continuous Features:} Discretize continuous features into bins, as entropy is typically computed for discrete distributions.

\textbf{Calculate Probability Distribution:} For each feature, calculate the probability distribution of the values or bins by counting the frequency of each value or bin and normalizing these counts to get probabilities.

\textbf{Compute Entropy:} Use the probability distribution to compute the entropy \( H(\phi(I)) \) of each feature \( \phi(I) \) using the formula:
\begin{equation}
H(\phi(I)) = -\sum_{i=1}^n P(x_i) \log P(x_i)
\end{equation}
where \( P(x_i) \) is the probability of the \( i \)-th value of feature \( \phi(I) \), and \( n \) is the total number of distinct values or bins.

\textbf{Compute Average Entropy:} Compute the average entropy for all vehicles by calculating the mean entropy across all features and all vehicles:
\begin{equation}
\bar{H} = \frac{1}{N} \sum_{j=1}^{N} H(\phi(I_j))
\end{equation}
where \( N \) is the number of unique vehicles and \( H(\phi(I_j)) \) is the entropy of the \( j \)-th vehicle.

\textbf{Analyze Consistency:} Low entropy indicates consistent feature values (less uncertainty), making them more reliable for distinguishing between different vehicles. High entropy indicates greater uncertainty andriability, suggesti vang the feature might be less reliable for vehicle re-identification.

\smallskip
\noindent
\textbf{Feature Consistency Analysis with Clustering Variance from T-SNE.}
Apply t-SNE to reduce the dimensionality of the features \textbf{F} to 2D vector \textbf{Z} for clustering results analysis and results visualization:
\begin{align}
\mathbf{Z} &= \text{t-SNE}(\mathbf{F})
\end{align}

where \( \mathbf{Z} \in \mathbb{R}^{n \times 2} \) are the 2D representations of the features.

Compute the variance of the t-SNE results for each condition:
\begin{equation}
    \text{Var}(\mathbf{Z}) = \left( \sigma_1^2, \sigma_2^2 \right)
\end{equation}
Calculate the RMSE (root mean square error) of the clustering variance:
\begin{equation}
    \text{RMSE} = \sqrt{\frac{1}{2} \sum_{i=1}^2 (\sigma_i^2 - \mu_i)^2}
\end{equation}
where \( \mu_i \) is the mean variance.

A higher RMSE value indicates that the feature distributions are more scattered, implying less consistency in the clustering. Conversely, a lower RMSE value indicates that the feature distributions are more compact, implying greater consistency.

\begin{table}[]
\centering
\caption{Vehicle data within each cameras.}
\label{tab:daa}
\resizebox{0.45\textwidth}{!}{%
\begin{tabular}{|c|c|c|c|}
\hline
\multirow{2}{*}{\textbf{Video Conditions}} & \multirow{2}{*}{\textbf{\#Vehicles}} & \textbf{InROI} & \textbf{OutROI} \\ \cline{3-4} 
                     &     & \textbf{\#Images} & \textbf{\#Images} \\ \hline
\textbf{sunny1}      & 73  & 635               & 912               \\ \hline
\textbf{sunny2}      & 132 & 1757              & 8020              \\ \hline
\textbf{rainy1}      & 64  & 794               & 2121              \\ \hline
\textbf{rainy2}      & 41  & 829               & 1364              \\ \hline
\textbf{night1}      & 26  & 320               & 490               \\ \hline
\textbf{night2}      & 6   & 30                & 82                \\ \hline
\textbf{congestion1} & 149 & 2496              & 3332              \\ \hline
\textbf{congestion2} & 52  & 777               & 2190              \\ \hline
\end{tabular}%
}
\end{table}

\begin{table}[]
\centering
\caption{Vehicle data cross cameras.}
\label{tab:daa2}
\resizebox{0.45\textwidth}{!}{%
\begin{tabular}{|c|c|c|c|}
\hline
\multirow{2}{*}{\textbf{Camera Pairs}} & \multirow{2}{*}{\textbf{\#Vehicles}} & \textbf{InROI} & \textbf{OutROI} \\ \cline{3-4} 
                        &    & \textbf{\#Images} & \textbf{\#Images} \\ \hline
\textbf{Pair 1 - cam 1} & 14 & 113               & -                 \\ \hline
\textbf{Pair 1 - cam 2} & 14 & 222               & 132               \\ \hline
\textbf{Pair 2 - cam 1} & 10 & 90                & -                 \\ \hline
\textbf{Pair 2 - cam 2} & 10 & 124               & 643               \\ \hline
\end{tabular}%
}
\end{table}

\begin{figure}
\centerline{\includegraphics[width=0.45\textwidth]{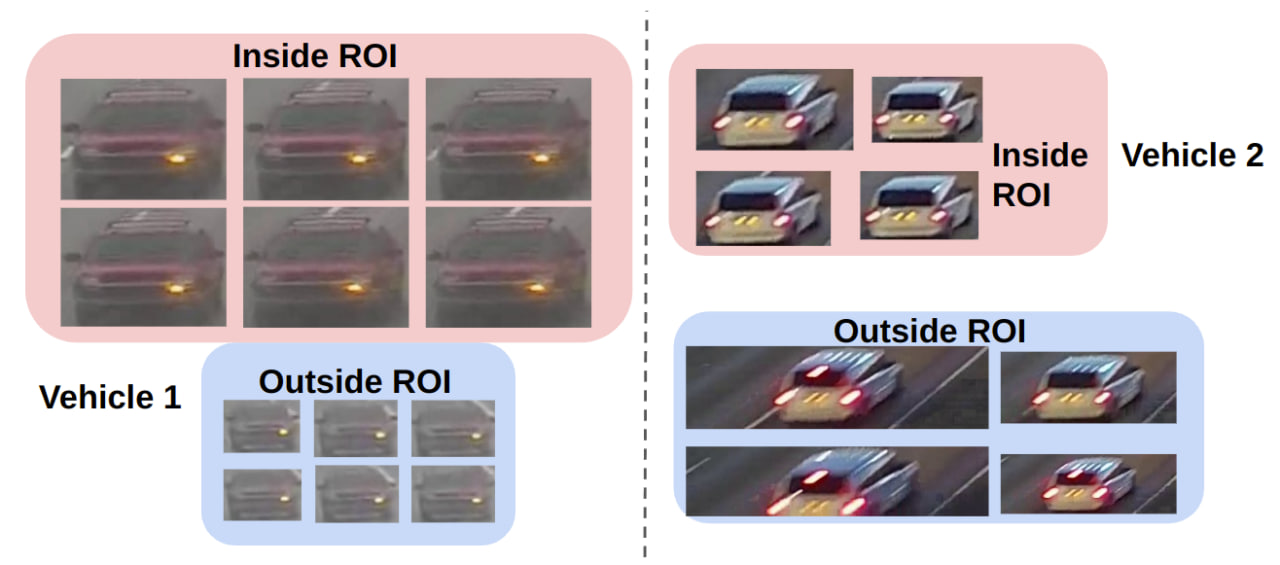}}
\vspace{-3mm}
\caption{\small \textit{Two vehicle examples from different camera views: For each vehicle, in-ROI and out-ROI images within the same camera view exhibit differences in size, aspect ratios, resolutions, and quality.}}
\label{veh-samples}
\vspace{-1mm}
\end{figure}

\begin{figure*}[h]
  \centering
  \subfigure(a){\includegraphics[width=0.22\textwidth]{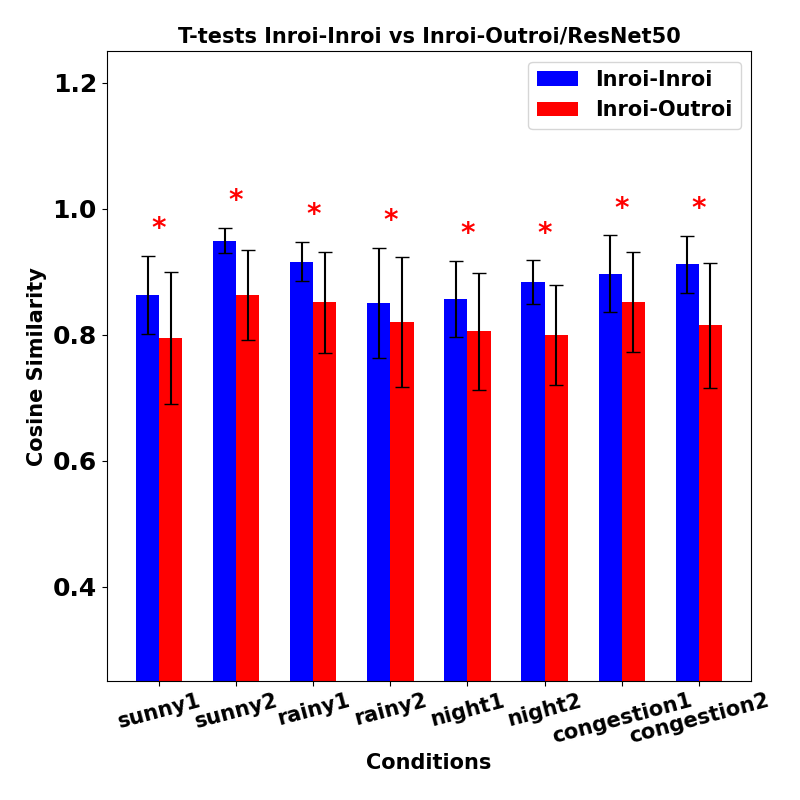}} 
  \subfigure(b){\includegraphics[width=0.22\textwidth]{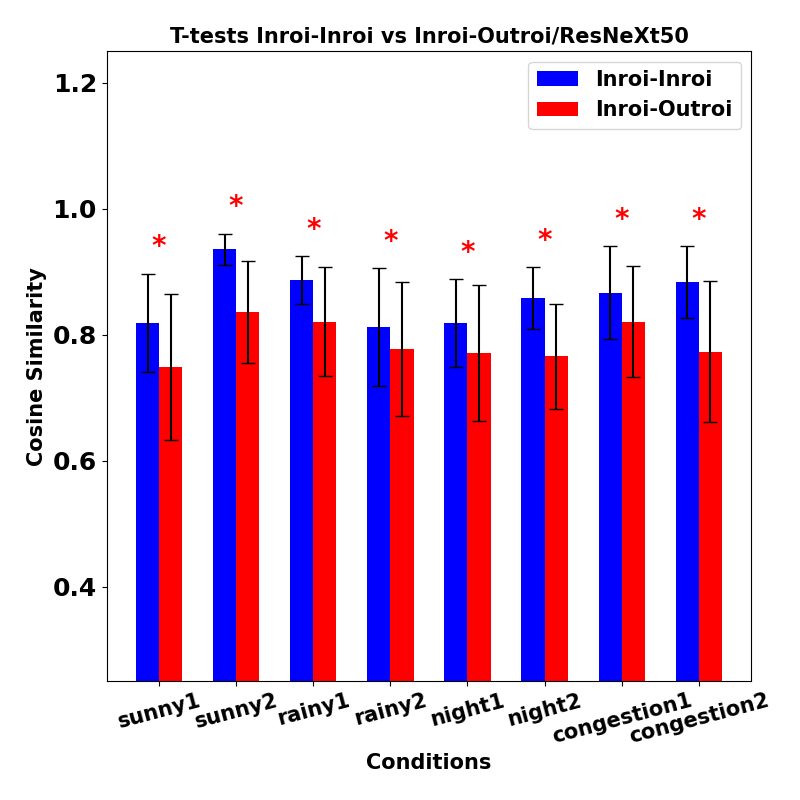}} 
  \subfigure(c){\includegraphics[width=0.22\textwidth]{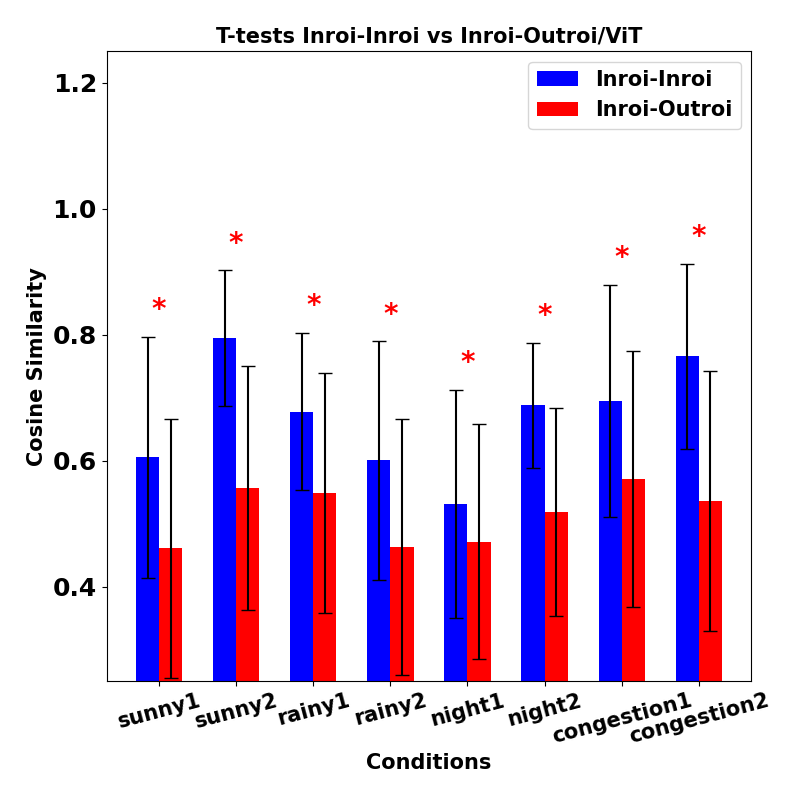}}
  \subfigure(d){\includegraphics[width=0.22\textwidth]{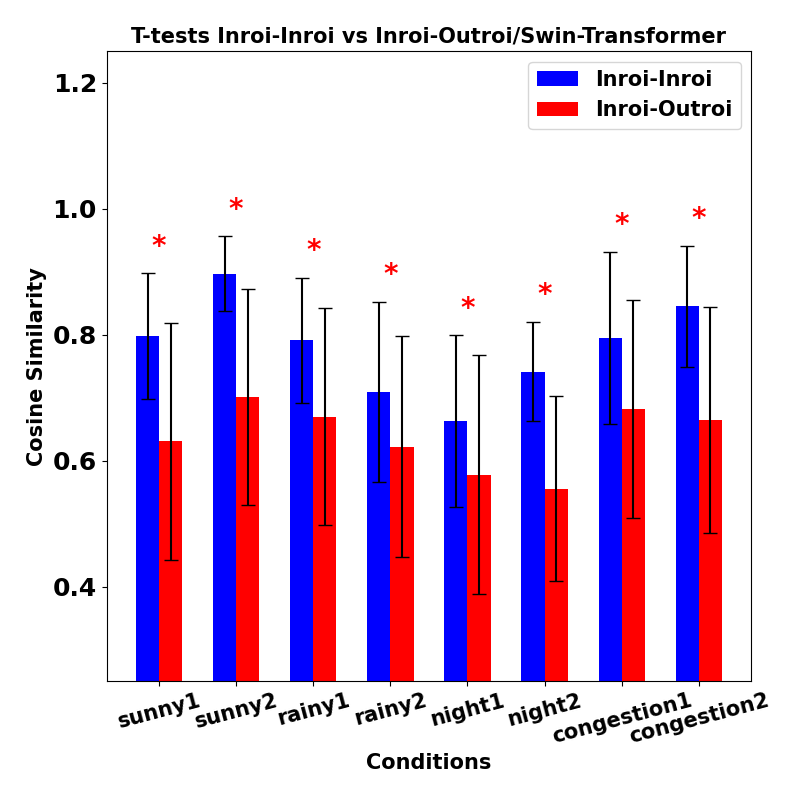}}
  \caption{\small \textit{Cosine similarity analysis was conducted using four pre-trained models (ResNet50, ResNeXt50, ViT, Swin-Transformer) across eight camera datasets under four conditions: sunny, rainy, night, and congestion. Features were extracted from the last layer before the fully connected (FC) layer. Cosine similarity was calculated for vehicle image pairs within the region of interest (in-ROI) and for pairs with one image in-ROI and the other out-ROI. T-tests were performed to determine significant differences in features between in-ROI and out-ROI, with a p-value threshold of 0.05. The T-test results from these models show significant differences between in-ROI and out-ROI features, with in-ROI vehicle features having higher similarities than those of out-ROI, as shown in (a), (b), (c), and (d). \textcolor{blue}{`*' means the T-test result is significant.}}}
\label{fig:t-test}
\end{figure*}

\begin{figure*}[h]
  \centering
  \subfigure(a){\includegraphics[width=0.45\textwidth]{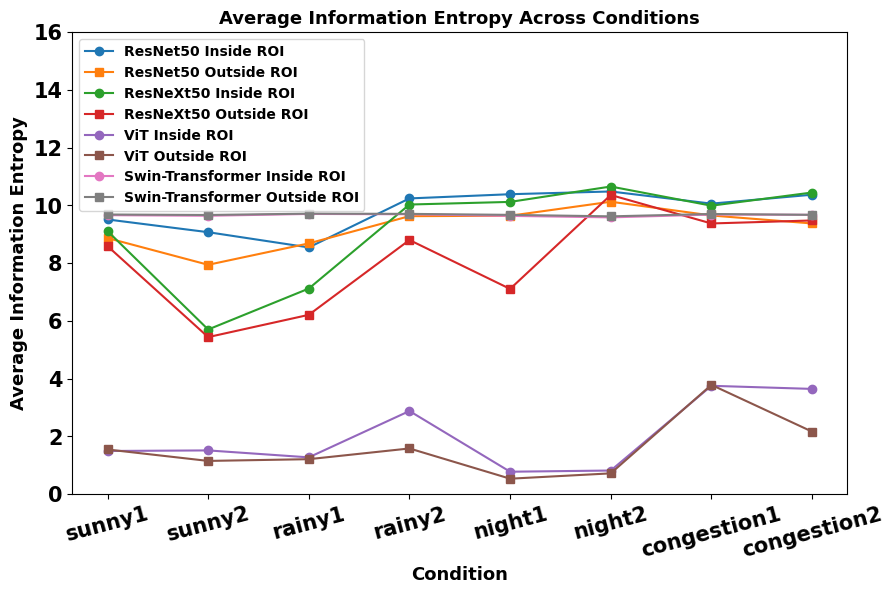}} 
  \subfigure(b){\includegraphics[width=0.45\textwidth]{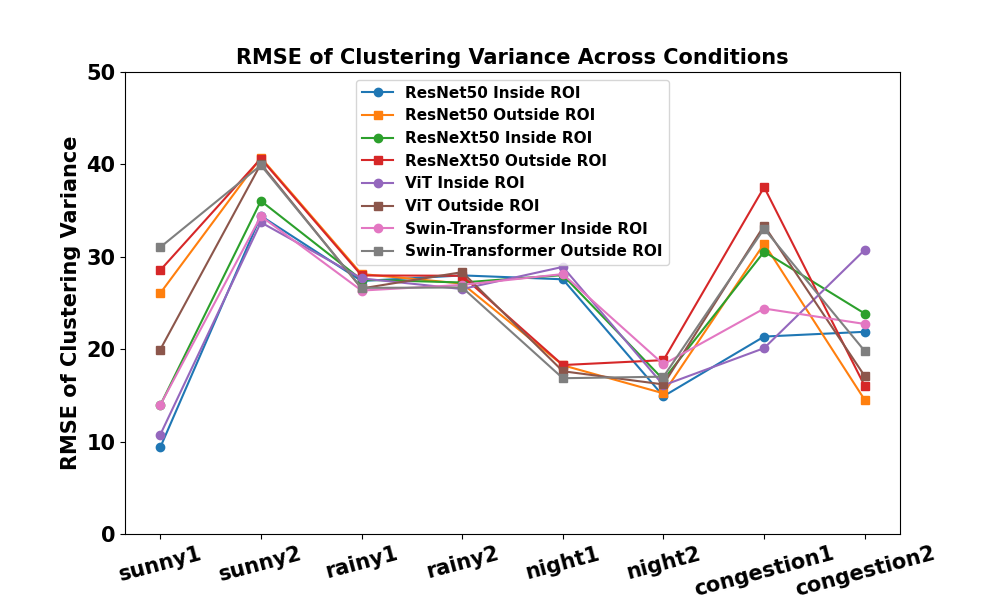}} 
\caption{\small \textit{
(a) Average Information Entropy Across Conditions: Analyzed within a single camera view, the average information entropy for in-ROI and out-ROI features from four models (ResNet50, ResNeXt50, ViT, and Swin-Transformer) across eight cameras and four conditions (sunny, rainy, night, and congestion). ViT exhibits the lowest entropy under all conditions, indicating robustness. Lower entropy signifies fewer variations in feature distribution. No significant difference in average entropy between in-ROI and out-ROI within each model. (b) RMSE of Clustering Variance Across Conditions: Post feature extraction, 2D t-SNE visualizations are generated, and clustering variance is calculated. Lower variance indicates more consistent features. For all four models, in-ROI features are more consistent than out-ROI features under challenging ITS conditions such as night and congestion.}}
\label{fig:info}
\end{figure*}

\section{Experiments}
\subsection{Experimental Settings}
\smallskip
\noindent
\textbf{Datasets.} 
The initial step involves collecting data from real ITS videos, utilizing 8 cameras covering various conditions: sunny, rainy, daytime, nighttime, and congested traffic. This setup includes 2 pairs of non-overlapping cameras, with one camera at the highway entry and the other at the highway exit. The data in TABLE \ref{tab:daa} and TABLE \ref{tab:daa2} provide an overview of vehicle counts and image distributions under various conditions and camera setups. TABLE \ref{tab:daa} details vehicle data across conditions: sunny1, sunny2, rainy1, rainy2, night1, night2, congestion1, and congestion2, including unique vehicles (\#Vehicles), InROI images, and OutROI images. For example, sunny1 recorded 73 vehicles with 635 InROI images and 912 OutROI images, while congestion1 recorded 149 vehicles with 2496 InROI images and 3332 OutROI images. TABLE \ref{tab:daa2} focuses on vehicle data from non-overlapping highway cameras, examining two pairs of cameras (Pair 1 and Pair 2). It lists the number of unique vehicles (\#Vehicles) and the corresponding InROI and OutROI image counts. For example, Pair 1 cam 1 captured 113 InROI images, while Pair 1 cam 2 captured 222 InROI images and 132 OutROI images. This data highlights differences in vehicle detection and image distribution, emphasizing the distinction between InROI and OutROI images. Fig. \ref{veh-samples} shows examples of vehicle images from different regions of the same camera view.
\begin{figure}[h]
  \centering
  \subfigure(a){\includegraphics[width=0.47\textwidth]{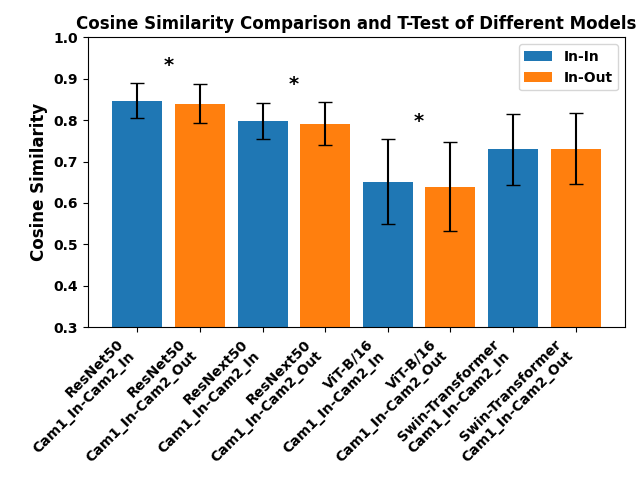}} 
  \subfigure(b){\includegraphics[width=0.47\textwidth]{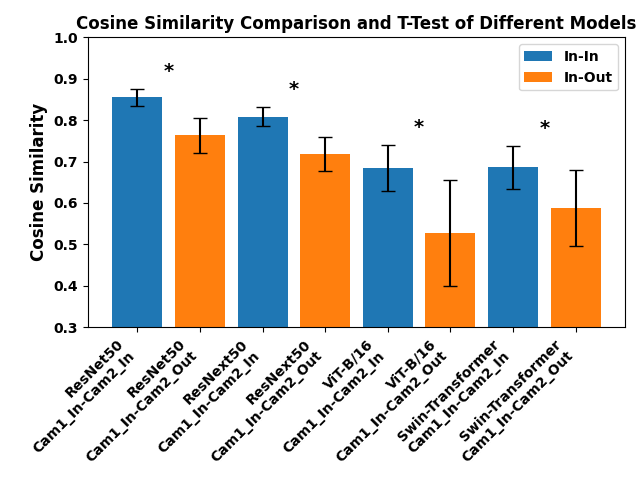}} \caption{\small \textit{\textbf{T-test of Cosine Similarity Across Cameras:} Cosine similarity was analyzed across four non-overlapping highway cameras: (a) the first pair of cameras, and (b) the second pair of cameras. The cosine similarity of the same vehicle's features was compared between cam1-inROI and cam2-inROI, and between cam1-inROI and cam2-outROI, using a p-value threshold of 0.05. Except for the first camera pair (a), features extracted by Swin-Transformer and other models showed significantly higher cosine similarity for inROI-inROI pairs compared to inROI-outROI pairs.\textcolor{blue}{`*' means the T-test result is significant.}}}
 
\label{fig:cross-cam}
\end{figure}

\begin{figure}[h]
  \centering
  \subfigure(a){\includegraphics[width=0.46\textwidth]{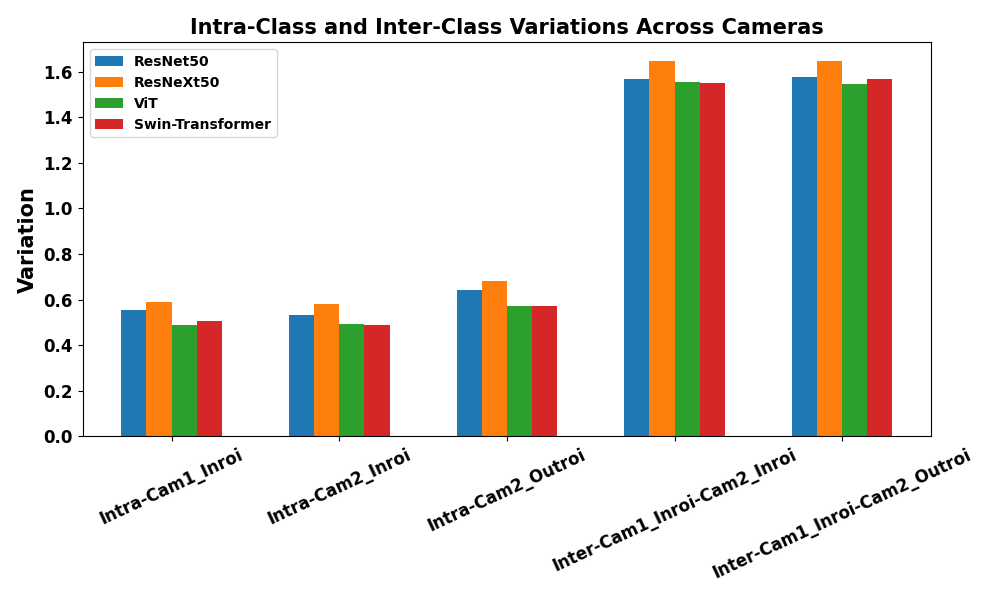}} 
  \subfigure(b){\includegraphics[width=0.46\textwidth]{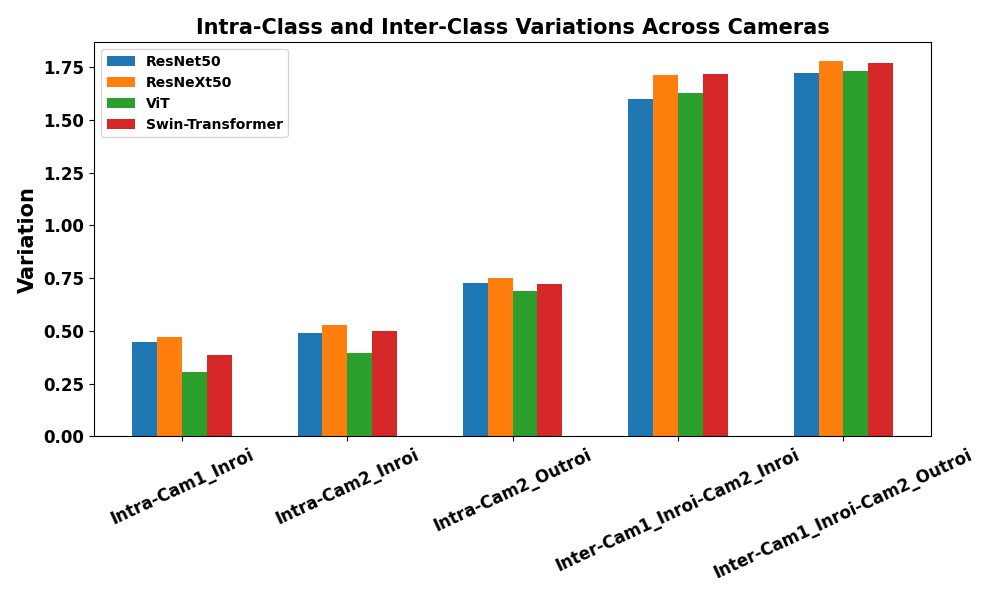}} 
\caption{\small \textit{\textbf{Intra-Class and Inter-Class Variations Across Cameras:} Clustering variance was analyzed across four non-overlapping highway cameras: (a) the first pair of cameras, and (b) the second pair of cameras. The intra-class feature variance is smaller than the inter-class feature variance in cross non-overlapping camera pairs. Additionally, the inter-class feature variance is smaller in ROI cross cameras compared to Outroi cross cameras. Among the models, ViT exhibits the smallest intra-class and inter-class variance values, indicating potentially higher robustness.}}
 
\label{fig:intra-inter}
\end{figure}
\smallskip
\noindent
\textbf{Implementation Details.} For data collection, we use YOLOv8 \cite{Jocher_Ultralytics_YOLO_2023} for vehicle detection and DeepSORT \cite{wojke2017simple} for tracking, utilizing their pre-trained weights. Ground truth within a single camera is generated by manually refining vehicle tracking results. For cross-camera scenarios, ground truth is generated by manually matching tracking results from corresponding camera pairs. For all four backbones, we load pre-trained models. All input images are resized to [224x224] and normalized with mean=[0.485, 0.456, 0.406] and std=[0.229, 0.224, 0.225]. Specifically, we use resnext50\_32x4d for ResNeXt50, vit\_base\_patch16\_224 for ViT, and swin\_base\_patch4\_window7\_224 for Swin-Transformer. No additional training is performed.



\begin{figure*}[h]
  \centering
  \subfigure(a){\includegraphics[width=0.45\textwidth]{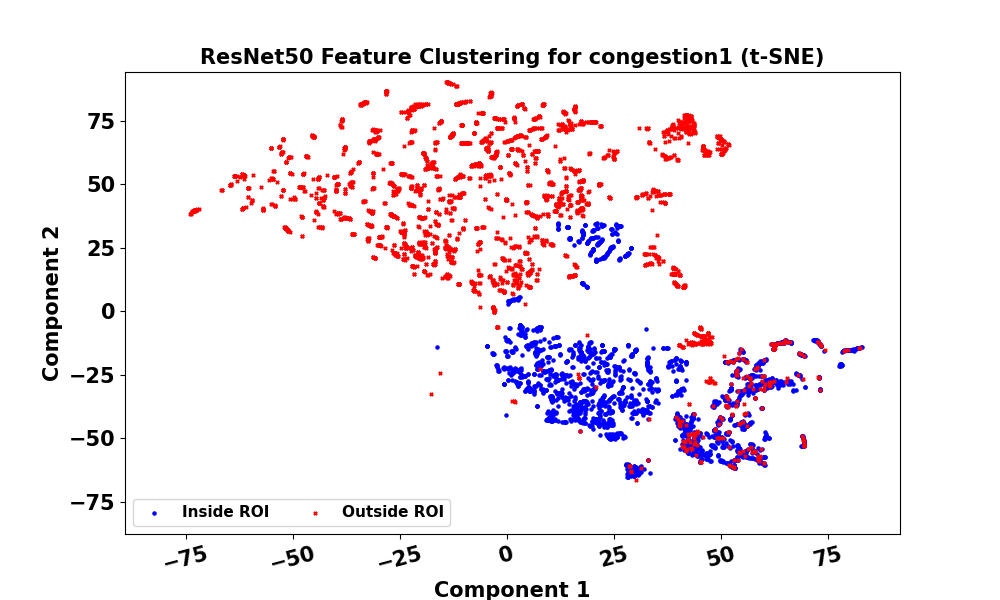}} 
  \subfigure(b){\includegraphics[width=0.45\textwidth]{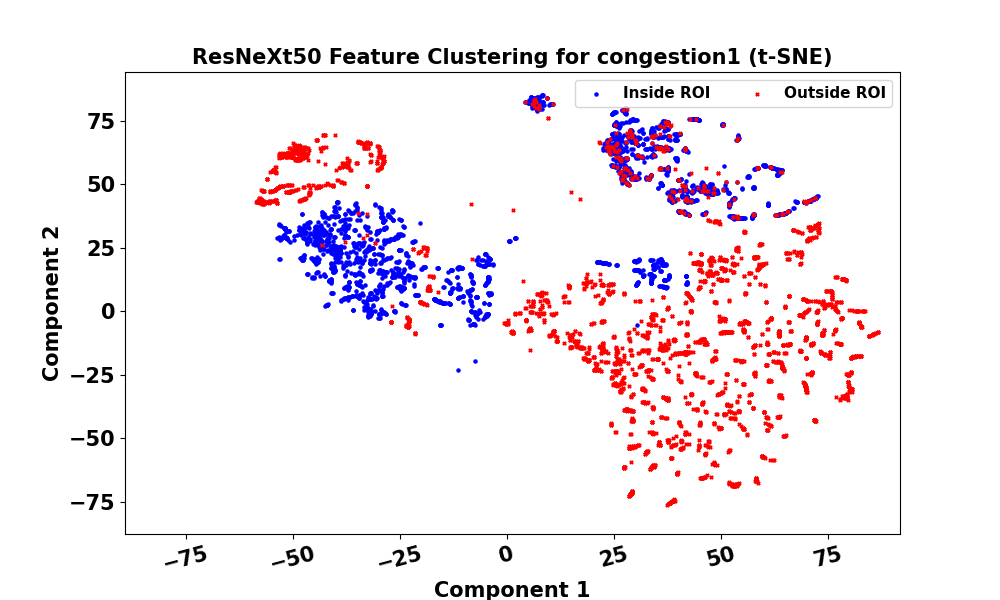}} 
  \subfigure(c){\includegraphics[width=0.45\textwidth]{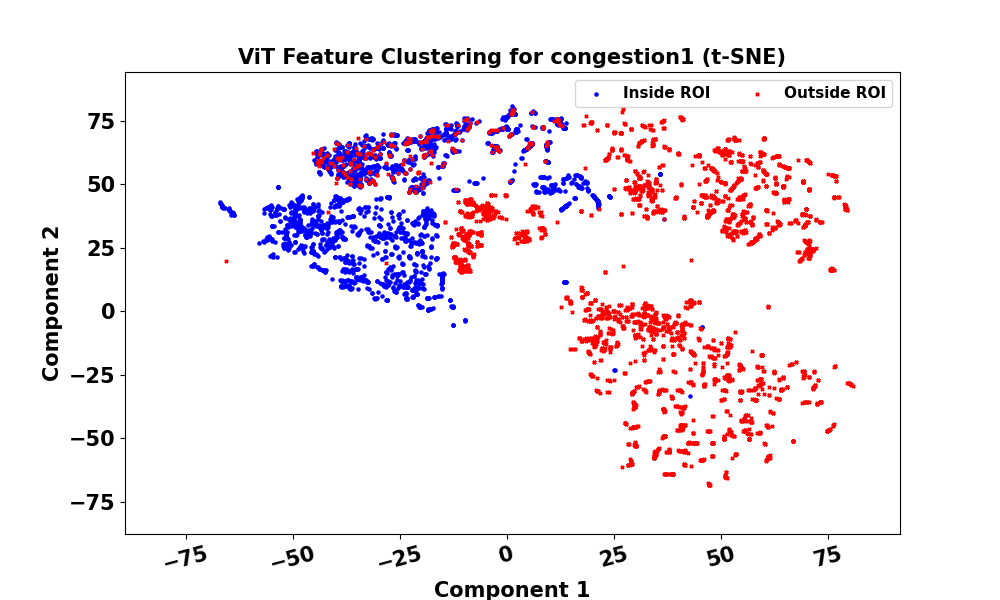}}
  \subfigure(d){\includegraphics[width=0.45\textwidth]{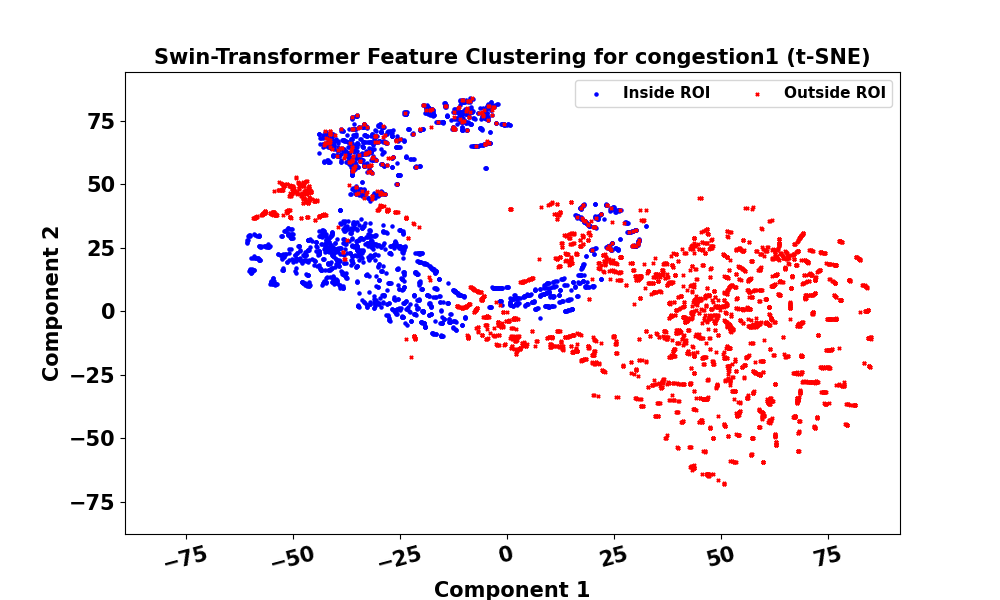}}
\caption{\small \textit{Within-camera, 8 cameras, 4 conditions: sunny, rainy, nighttime, and congestion; 2D t-SNE clustering of features extracted from four models (ResNet50, ResNeXt50, ViT, Swin-Transformer) for both in-ROI and out-ROI. Features from out-ROI are more scattered than in-ROI, indicating higher variation in out-ROI features. (a) ResNet50 Feature Clustering for congestion1, (b) ResNeXt50 Feature Clustering for congestion1, (c) ViT Feature Clustering for congestion1, (d) Swin-Transformer Feature Clustering for congestion1. The blue dots indicate vehicles inside ROI, and the red dots indicate vehicles outside ROI. Here, we only set two classes: inroi and outroi, not considering any specific vehicle IDs.}}
 
\label{fig:t-sne}
\end{figure*}

\subsection{Results}
\smallskip
\noindent
\textbf{Within Cameras.}
Within a single camera view, our T-test results of cosine similarity across eight cases reveal a significant difference between InROI and OutROI, as depicted in Fig. \ref{fig:t-test} for all four backbones. Average information entropy and RMSE of clustering variance from t-SNE results show that InROI features are more consistent than OutROI features. Furthermore, the Vision Transformer (ViT) demonstrates superior robustness in challenging ITS scenarios, as shown in Fig. \ref{fig:info}. To visualize it, the clustering results of one case are shown in Fig. \ref{fig:t-sne}.

\smallskip
\noindent
\textbf{Cross non-overlapping Cameras.}
The T-test results of the four models on cross-camera data are presented in Fig. \ref{fig:cross-cam}. Except for the first camera pair shown in Fig. \ref{fig:cross-cam}(a), features extracted by the Swin-Transformer and other models from InROI regions of both cameras exhibit significantly higher cosine similarity compared to those from the InROI of one camera and the OutROI of the other. Fig. \ref{fig:intra-inter} shows that intra-class feature variance is smaller than inter-class feature variance across non-overlapping camera pairs. Additionally, inter-class feature variance is lower for InROI cross-camera comparisons than for OutROI cross-camera comparisons. Among the models, the Vision Transformer (ViT) demonstrates the smallest intra-class and inter-class variance values, indicating potentially higher robustness.

\section{Conclusion}
We implemented our previous framework, integrating automatic optimal ROI learning, real-time vehicle detection, and tracking on eight highway camera videos under four ITS conditions: sunny, rainy, nighttime, and congestion, plus two pairs of non-overlapping cameras. Features were extracted using four classical backbones. Cosine similarity, information entropy, and T-SNE clustering variance were used to analyze these features. Our results show significant differences in feature consistency between ROIs and non-ROIs, suggesting that optimizing ROIs based on vehicle detection results can enhance vehicle ReID in ITS. Further research is needed due to the limited dataset size.

\printbibliography

\end{document}